\documentclass[letterpaper, 10 pt, conference]{ieeeconf}  

\IEEEoverridecommandlockouts                              
\usepackage{url}
\usepackage{amsmath,amssymb,bm}
\usepackage{graphicx}
\usepackage{algorithm}
\usepackage{algpseudocode}
\usepackage{xcolor}
\usepackage{booktabs}
\usepackage{array}
\usepackage{bbm}
\usepackage{hyperref}
\usepackage{xurl}

\title{\LARGE \bf
Module-Number-Adaptive Visual Shape Control for Serial Modular Soft Robots
}
\author{Kyohei Akamine$^{1}$, Takato Horii$^{1}$, Yusuke Sakaue$^{1}$, and Hiroki Ishizuka$^{2}$%
\thanks{This study was supported by JST PRESTO Grant Number JPMJPR22S2, JST CREST Grant Number JPMJCR2555, and JSPS KAKENHI Grant numbers 22H01447 and 23H01379, Japan.}%
\thanks{$^{1}$K. Akamine, T. Horii, and Y. Sakaue are with the Graduate School of Engineering Science, The University of Osaka, Osaka 560-0043, Japan (e-mail: akamine@bpe.es.osaka-u.ac.jp; takato@sys.es.osaka-u.ac.jp; sakaue.yuusuke.es@osaka-u.ac.jp).}%
\thanks{$^{2}$H. Ishizuka is with the Faculty of Science and Technology, Sophia University, 7-1 Kioi-cho, Chiyoda-ku, Tokyo 102-8554, Japan (e-mail: hiroki.ishizuka.soro@gmail.com).}}

\begin{document}

\maketitle
\thispagestyle{empty}
\pagestyle{empty}

\begin{abstract}
Image-based shape control provides a simple means of controlling the whole-body configuration of soft robots. However, existing data-driven approaches are typically developed for fixed robot structures and require new control data when the number of modules changes. This paper presents a module-number-adaptive visual shape control method for serial modular soft pneumatic robots. A controller trained only on single-module actuation–shape data is reused for robots with one to five modules by decomposing whole-body camera images into local module patches. 
A single common module segmenter localizes individual modules across all tested configurations, while the same local controller is applied to every extracted patch.
Geometric data augmentation improves transferability to downstream modules, and a lightweight mask reconstruction network reconstructs a synthetically removed actuator-mask channel. Experiments on physical robots demonstrate shape control across varying numbers of modules and under environmental changes and payload loading. The results show that single-module control learning enables scalable whole-body control without configuration-specific control-data collection.
Code is available at \url{https://anonymous.4open.science/r/3D-SMR-Controller-9574}.

\end{abstract}

\section{INTRODUCTION}
In recent years, soft robots composed of flexible materials and deformable structures have attracted considerable attention for applications in environments that are difficult for conventional rigid robots, including exploration in confined spaces, medical assistance, environmental monitoring, and object manipulation. Among soft robotic systems, modular soft pneumatic robots constructed by connecting multiple deformable modules can generate a wide variety of postures and motions by combining the deformation of individual modules, thereby achieving behaviors that cannot be realized by a single module~\cite{tang2026review}.

Accurate analytical modeling of soft robots is difficult because their behavior is affected by high degrees of freedom, material nonlinearities, hysteresis, external-force interactions, and manufacturing variability~\cite{shariati2021, Ferrentino2022, kosaka2025}. Model-based control methods using curvature models, Cosserat rod models, and finite-element models have therefore been widely investigated~\cite{Webster2010, DellaSantina2020, DellaSantinaSurvey2023, Till2019, Largilliere2015, Tonkens2021}. In physical systems, however, discrepancies often arise between model assumptions and actual robot behavior. This problem is particularly pronounced in modular soft robots, because a control mapping must first be learned for modules composed of nonlinear soft actuators, and these modules must subsequently be coordinated when connected into a larger structure. Conventional
data-driven methods typically learn this mapping for the entire robot,
making the learned controller difficult to reuse when modules are added or removed.

As an alternative, data-driven control methods based on deep learning and reinforcement learning have attracted increasing attention~\cite{Chen2025,ChinMajidi2020}. These methods learn the relationship between observations and control inputs directly from physical robot data, allowing complex behaviors involving nonlinearities and robot-specific variations to be handled without detailed geometric or mechanical models~\cite{ABCD}. In modular soft robots, such approaches may also provide a means of learning nonlinear behavior and inter-module interactions associated with structural changes.

In this study, we focus on whole-body shape control of soft robots using image targets representing the entire shape~\cite{Elijah2023, Marques2024}. Because the entire body of a soft robot deforms continuously, controlling only the end-effector position is insufficient for safe operation in confined spaces or near obstacles; the whole-body shape must also be considered as a control objective. 
Image-based control methods have traditionally focused on the end-effector
positioning, although recent studies have extended visual servoing to
whole-body shape control of multi-section continuum robots~\cite{gandhi20263dshapecontrolextensible}. However, because redundant soft robots can assume multiple body configurations for the same tip position, such methods make it difficult to explicitly regulate intermediate body shapes and interactions with the environment.
Almanzor et al. proposed an image-based whole-body control method in which the current and target shapes are represented in the same image space~\cite{Elijah2023}. Camera images provide distributed, non-contact observations of robot deformation without requiring expensive motion-capture systems or embedded sensors~\cite{usui2021, soter2019}.
Nevertheless, existing image-based whole-body control methods have primarily been developed for fixed-structure soft continuum robots, and their application to robots with varying numbers of modules remains underexplored. 
This issue is particularly important in image-based whole-body shape control, where a target image represents the desired configuration of the entire robot. Even when the robot is assembled from identical modules, each module occupies a different position and orientation within the target image after the modules are connected. These configuration-dependent visual representations make it difficult to reuse a control model learned from a single module across different module positions and module counts.

Control of modular soft robots has also been investigated using sequence-based models that explicitly represent module states along a serial chain. Chen et al. employed a bidirectional long short-term memory network together with state measurements obtained from multiple cameras and optical markers~\cite{chen2024}.
Their method addresses the coordination of connected modules using explicitly measured module states, rather than whole-body shape control
in which the desired robot configuration is specified by a target image. Moreover, the controller was learned using a three-module structure, and the reuse of control knowledge learned from a single module was not the focus of the study.


In this study, we propose a shape control
method for serial pneumatic soft robots composed of identical soft modules using image targets. A local
controller is trained using only single-module visual and actuation
data, and it is then reused across multi-module configurations without
additional controller training. 
The controller requires no configuration-specific control data or retraining, while a single common module segmenter is fine-tuned on a combined dataset containing images from all tested configurations.
Whole-body control is achieved by
decomposing the current and target robot images into module-level
patches and applying the same controller to each module.

When identical modules are connected in series, their local image
representations vary according to their positions in the robot.
In particular, downstream modules appear at different image
locations, orientations, and scales from those observed in the
single-module training data. To reduce this position-dependent visual
distribution shift, geometric data augmentation is applied during
single-module training. This allows the learned local visual control
mapping to be reused across different module positions and 
counts.
We also evaluate the transferred controller not only across different module numbers, but
also under lighting changes, background disturbances, and a tip payload.

Posture changes, module overlap, and partial occlusion may cause parts
of the module masks to become unavailable. We therefore introduce a
lightweight mask reconstruction network that restores missing visual
features and provides a consistent image representation to the local
controller. The proposed framework consequently enables markerless
whole-body shape control using a single camera while reusing control
knowledge learned from a single module.

Table~\ref{tab:comparison} compares the proposed method with existing
image-based control approaches. Unlike prior methods, the proposed
framework focuses on the compositional reuse of a local controller
trained only from single-module control data across serial
configurations with different numbers of identical modules. The
contributions of this study are summarized as follows.

\begin{itemize}
\item A compositional visual control framework that reuses a single-module controller.
\item Transfer of a controller trained from single-module control data to
one- to five-module configurations without controller retraining.
\item Experimental validation on physical robots with different module numbers.
\end{itemize}

\begin{table}[t]
  \centering
  \caption{Comparison with prior work.}
  \label{tab:comparison}
  \renewcommand{\arraystretch}{1.15}
  \setlength{\tabcolsep}{2pt}
  \resizebox{\columnwidth}{!}{%
  \begin{tabular}{lcccc}
  \toprule
  \textbf{Method} &
  \textbf{Sensor} &
  \textbf{Scale} &
  \textbf{Feedback} &
  \textbf{VC} \\
  \midrule

  Almanzor \textit{et al.}~\cite{Elijah2023} &
  RGB &
  Fixed &
  Visual &
  $\times$ \\

  Monteiro \textit{et al.}~\cite{Marques2024} &
  RGB &
  Fixed &
  Open-loop &
  $\times$ \\

  Chen \textit{et al.}~\cite{chen2024} &
  MoCap+markers &
  $3{\to}2,4$ &
  Configuration &
  $\times$ \\

  \textbf{Ours} &
  RGB &
  $1{\to}5$ &
  Visual &
  $\checkmark$ \\

  \bottomrule
  \end{tabular}%
  }

  \vspace{1mm}
  \scriptsize
  RGB: single RGB camera; MoCap: Vicon optical motion capture;
  VC: visual completion. Scale denotes the module number used for controller training followed by the module numbers used for physical deployment.
\end{table}

\section{SOFT ROBOT DESIGN}

\begin{figure}[t]
  \centering
  \includegraphics[width=1.0\linewidth]{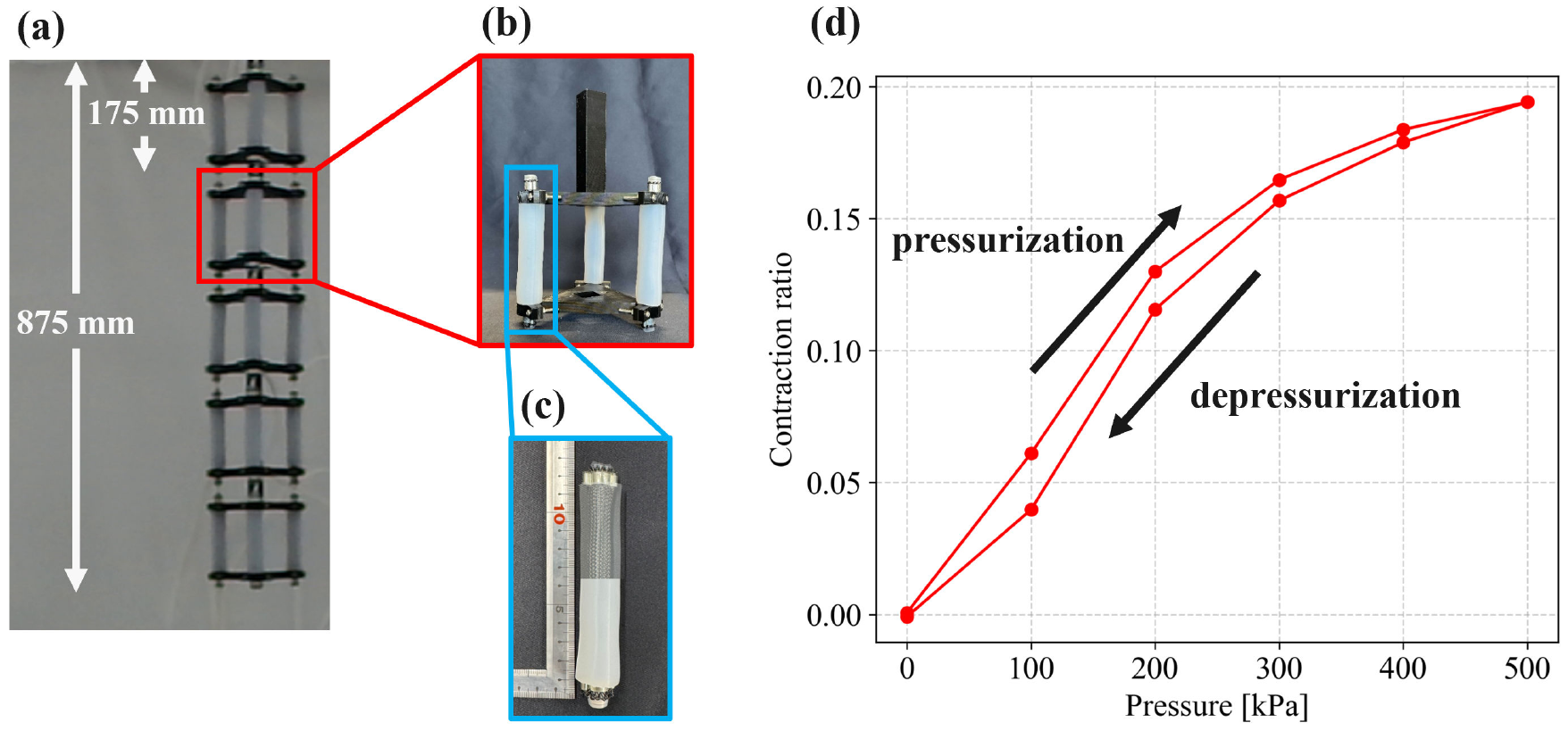}
  \caption{A soft robot is constructed by connecting modules, each consisting of three soft pneumatic actuators. (a) Five-module soft robot. Its total length is 875~mm, and it can bend up to 75$^\circ$ in each direction. (b), (c) A module and its constituent pneumatic artificial muscle actuators. (d) Pressure--contraction curve. The actuator contracts by up to approximately 20\%.}
  \label{fig:smr}
\end{figure}

To evaluate the proposed controller, we fabricated a modular soft pneumatic robot, as shown in Fig.~\ref{fig:smr}(a)--(c). Each module comprises three McKibben-type pneumatic artificial muscles~\cite{recipe} arranged at the vertices of an equilateral triangular frame. Their independent contraction enables three-dimensional deformation. 
To reduce buckling caused by self-weight, each muscle was enclosed in a 5-mm-thick cylindrical silicone skin (Ecoflex 00-50 Supersoft, Smooth-On). As shown in Fig.~\ref{fig:smr}(d), the skinned actuator achieved a maximum contraction ratio of approximately 20\% over the operating pressure range of 0--500 kPa.
A single module is approximately 175 mm long and bends up to 15$^\circ$. Five serially connected modules form an approximately 875-mm-long robot with a maximum cumulative bending angle of 75$^\circ$.

\section{PROPOSED CONTROL METHOD}
\begin{figure}[t]
  \centering
  \includegraphics[width=1.0\linewidth]{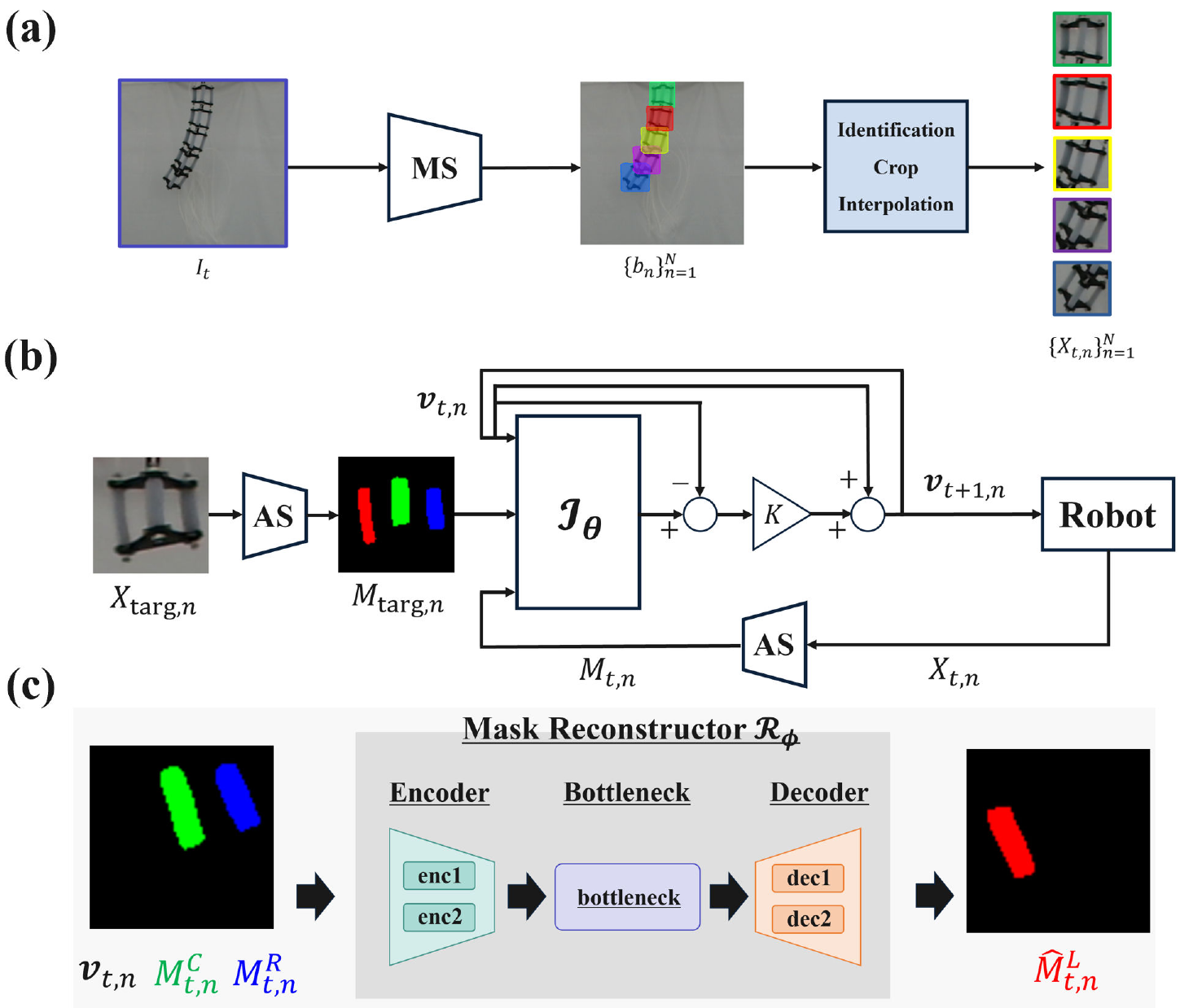}
  \caption{Architecture of the proposed control model, which deploys a module controller trained on a single module to multi-module configurations without additional multi-module control-data collection or controller retraining. (a) Module extractor using the module segmenter (MS) to extract $N$ module patches. (b) Module controller using the actuator segmenter (AS) and voltage prediction model $\mathcal{I}_\theta$. (c) Mask reconstruction network $\mathcal{R}_\phi$ that reconstructs a missing actuator-mask channel.}
  \label{fig:method}
\end{figure}

This section describes the proposed control model shown in Fig.~\ref{fig:method}. The model consists of a module extractor, shown in Fig.~\ref{fig:method}(a), a module controller, shown in Fig.~\ref{fig:method}(b), and a mask reconstruction network, shown in Fig.~\ref{fig:method}(c). 
Let $N$ denote the number of modules detected in the whole-body image $I_t$ of the robot at control step $t$. The module extractor generates a patch $X_{t,n}$ corresponding to the $n$th module, where $n=1,\ldots,N$. The current patch is represented by the three-channel binary mask $M_{t,n}=[M^L_{t,n},M^C_{t,n},M^R_{t,n}]$, whose channels correspond to the left, center, and right artificial muscles, respectively. The target patch is represented analogously by $M_{\mathrm{targ},n}=[M^L_{\mathrm{targ},n},M^C_{\mathrm{targ},n},M^R_{\mathrm{targ},n}]$. Let $\boldsymbol{v}_{t,n}=[v^L_{t,n},v^C_{t,n},v^R_{t,n}]^\top\in[0,V_{\max}]^3$ denote the command voltages applied to these muscles at control step $t$, where $V_{\max}=5.0$~V. Based on $M_{t,n}$, $M_{\mathrm{targ},n}$, and $\boldsymbol{v}_{t,n}$, the module controller computes the next-step command voltage $\boldsymbol{v}_{t+1,n}\in[0,V_{\max}]^3$. 
For estimating the module regions and actuator shapes, we use Detectron2~\cite{wu2019detectron2} to construct and fine-tune two instance-segmentation models on separate custom datasets: a module segmenter (MS) and an actuator segmenter (AS), respectively.
When a mask is missing, the reconstruction network shown in Fig.~\ref{fig:method}(c) restores the missing mask.

The module extractor localizes modules in each configuration,
while the module controller addresses control scalability by applying the
same controller to every detected module patch. By decomposing the whole-body image into local module-level patches and independently estimating the control input for each patch, the proposed architecture enables a control model trained on a single module to be deployed to multi-module configurations without additional controller training.

The module controller employs a closed-loop structure that iteratively updates the applied voltages according to the difference between the current and target states. 
Furthermore, the mask reconstruction network maintains a consistent representation of the controller input when visual information is partially lost because of occlusion or overlap between modules.

\subsection{Module Extractor}

\subsubsection{Algorithm}
The module segmenter (MS) is applied to the entire image $I_t$ to identify the module regions and their bounding boxes $\{\boldsymbol{b}_n\}_{n=1}^{N}$, where $\boldsymbol{b}_n=(x_{1,n},y_{1,n},x_{2,n},y_{2,n})$ denotes a detected module bounding box. For each box, its center $y$-coordinate is computed as $c_n^y=(y_{1,n}+y_{2,n})/2$, and the boxes are sorted in ascending order of $c_n^y$. The module index $n$ is thereby assigned from the top to the bottom of the robot, providing a consistent correspondence between the current and target images. Next, each corresponding region is cropped from $I_t$. An interpolation process is then applied so that the cropped region smoothly transitions to the surrounding background, yielding a final patch with reduced background noise. By performing this process for all bounding boxes, the patch set $\{X_{t,n}\}_{n=1}^{N}$ is obtained.

\subsubsection{Training}
The module segmenter, which treats the entire module as a single object class, was initialized with COCO-pretrained weights and fine-tuned using a custom annotated dataset consisting of 50 randomly posed images from each of the one- to five-module configurations. The same segmenter was used for all experiments, and no module-number-specific segmenter was trained.
The images were manually annotated using LabelMe~\cite{wada2021labelme}. Data collection required approximately 5 min, and the annotation and fine-tuning process required approximately 30 min.

\subsection{Module Controller}
\subsubsection{Algorithm}
First, the actuator segmenter (AS) is applied to each patch $X_{t,n}$ extracted by the module extractor. Let $\{(m_j,s_j)\}_{j=1}^{J}$ denote the candidate binary masks and confidence scores returned across the three actuator classes, where $J$ is the number of candidate instances returned by AS. Masks with extremely small areas or confidence scores below a predefined threshold are discarded as noise, after which the three highest-scoring masks are retained. For each retained mask, let $\Omega_j=\{(u,v)\mid m_j(u,v)=1\}$ denote its foreground region, and let ${c}_j^x=|\Omega_j|^{-1}\sum_{(u,v)\in\Omega_j}u$ denote its centroid $x$-coordinate. Although AS predicts three actuator classes, the retained masks are reordered according to ${c}_j^x$ to enforce a consistent spatial channel order independent of the segmenter output order. The masks are assigned from left to right and combined to construct $M_{t,n}$. The target whole-body image $I_{\mathrm{targ}}$ is processed in advance by the module extractor using the same procedure applied to $I_t$, yielding the target patch $X_{\mathrm{targ},n}$. The target mask $M_{\mathrm{targ},n}$ is then computed from $X_{\mathrm{targ},n}$ using the same actuator-segmentation procedure.

Next, the voltage vector is normalized by $V_{\max}$ and broadcast over the spatial dimensions of the patch to form the three-channel voltage map $Q_{t,n}=\operatorname{Broadcast}(\boldsymbol{v}_{t,n}/V_{\max})$. Thus, every spatial element in each channel of $Q_{t,n}$ contains the normalized command voltage of the corresponding artificial muscle. The current mask, target mask, and voltage map are concatenated to form the nine-channel input tensor $X^{\mathrm{CNN}}_{t,n}$ for the CNN-based voltage prediction model $\mathcal{I}_\theta$, where $\theta$ denotes its learned parameters.

The voltage prediction model consists of a CNN encoder and an MLP regression head. The CNN encoder comprises three successive convolution--BatchNorm--ReLU--MaxPool blocks followed by adaptive average pooling, while the regression head consists of two fully connected layers with dropout. From the nine-channel input tensor, the model estimates the desired three-dimensional command voltage $\hat{\boldsymbol{v}}_{t,n}$. A first-order update with gain $K=0.5$ is then applied to the difference between $\hat{\boldsymbol{v}}_{t,n}$ and the current voltage $\boldsymbol{v}_{t,n}$, yielding the unclipped next-step voltage $\boldsymbol{v}^{\mathrm{raw}}_{t+1,n}$. Finally, this voltage is clipped to $[0,V_{\max}]^3$ to obtain $\boldsymbol{v}_{t+1,n}$.

\begin{algorithm}[h]
\caption{Module Controller}
\label{alg:controller}
\begin{algorithmic}[1]
\Require $X_{t,n}$, $X_{\mathrm{targ},n}$, $\boldsymbol{v}_{t,n}$, $K=0.5$, $V_{\max}=5.0$~V
\Ensure $\boldsymbol{v}_{t+1,n}$
\State $\{(m_j,s_j)\}_{j=1}^{J} \leftarrow \mathrm{AS}(X_{t,n})$
  \Comment{Estimate candidate masks and scores across three classes}
\State Discard masks below the score and area thresholds; retain the top three
\State $\Omega_j \leftarrow \{(u,v)\mid m_j(u,v)=1\}$ for each retained mask
\State ${c}_j^x \leftarrow \frac{1}{|\Omega_j|}\sum_{(u,v)\in\Omega_j} u$
  \Comment{Compute centroid $x$-coordinate}
\State $M_{t,n} \leftarrow \mathrm{sort}(m_j,\ c_j^x,\ \text{ascending})$
  \Comment{Canonicalize the spatial order as left, center, right}
\State $M_{\mathrm{targ},n} \leftarrow$ the same mask-extraction procedure applied to $X_{\mathrm{targ},n}$
\State $Q_{t,n} \leftarrow \operatorname{Broadcast}(\boldsymbol{v}_{t,n} / V_{\max})$
  \Comment{Normalize and generate voltage map}
\State $X^{\mathrm{CNN}}_{t,n} \leftarrow \mathrm{concat}(M_{t,n},\, M_{\mathrm{targ},n},\, Q_{t,n})$
\State $\hat{\boldsymbol{v}}_{t,n} \leftarrow \mathcal{I}_\theta(X^{\mathrm{CNN}}_{t,n})$
  \Comment{Estimate voltage}
\State $\boldsymbol{v}^{\mathrm{raw}}_{t+1,n} \leftarrow \boldsymbol{v}_{t,n} + K(\hat{\boldsymbol{v}}_{t,n} - \boldsymbol{v}_{t,n})$
  \Comment{Update voltage}
\State $\boldsymbol{v}_{t+1,n} \leftarrow \mathrm{clip}(\boldsymbol{v}^{\mathrm{raw}}_{t+1,n},\ 0,\ V_{\max})$
  \Comment{Clipping}
\State \Return $\boldsymbol{v}_{t+1,n}$
\end{algorithmic}
\end{algorithm}

\subsubsection{Training}
The actuator segmenter, trained with three classes—left, center, and right—was initialized with COCO-pretrained weights and fine-tuned using a custom annotated dataset consisting of 50 images extracted by the module extractor and 50 randomly posed images of the single-module configuration. The annotation and fine-tuning process required approximately 1 h.

To train the voltage prediction model, voltages ranging from 0.0 to 5.0 V in increments of 0.5 V were applied independently to the three artificial muscles of the single-module robot, yielding 1331 ($=11^3$) voltage--mask image pairs. To emulate the apparent displacement of downstream modules in multi-module configurations, each sample was augmented 20 times using random rotations and translations.

During training, two samples—referred to as the anchor and the partner—were randomly selected from the training dataset. The anchor voltage map, the anchor mask, and the partner mask were used as inputs, while the voltage associated with the partner sample was used as the supervisory target. The model was optimized using Adam to minimize the Smooth L1 loss. Data collection required approximately 25 min, and model training required approximately 4 h.

\subsection{Mask Reconstruction Network}

\subsubsection{Algorithm}
In mask estimation, a specific mask channel may be missing due to changes in visual conditions, such as illumination variation or overlap between adjacent muscles. Failure to estimate the required number of masks can hinder proper control of the soft robot. To address this issue, we introduce a lightweight U-Net-based reconstruction network $\mathcal{R}_\phi$, where $\phi$ denotes its learned parameters. The same procedure can be applied to any missing channel by using the other two channels as inputs and the missing channel as the training target. Here, we describe the left-channel case as a representative example. As illustrated in Fig.~\ref{fig:method}(c), the external inputs are the non-missing center and right mask channels, $M^C_{t,n}$ and $M^R_{t,n}$, and the current voltage vector $\boldsymbol{v}_{t,n}$. The voltage vector is normalized and broadcast to form $Q_{t,n}$ as defined above, and $\mathcal{R}_\phi$ takes the concatenation of $M^C_{t,n}$, $M^R_{t,n}$, and $Q_{t,n}$ as its five-channel input. The network outputs the left-mask logit map $\hat{L}_{t,n}$. Applying the sigmoid function $\sigma(\cdot)$ converts this logit map into a pixelwise foreground-probability map. At inference time, pixels with probabilities greater than 0.5 are assigned a value of one and the remaining pixels are assigned zero, yielding the reconstructed binary left mask $\hat{M}^L_{t,n}$.

\subsubsection{Training}
The reconstruction network was trained using the same 1331 base voltage--mask image pairs employed to train the voltage prediction model of the module controller, augmented in the same manner. For each sample, the input was a five-channel tensor formed by concatenating the two-channel center and right masks with the three-channel voltage map, while the corresponding left mask was used as the ground-truth target. The loss function was defined as a sum of the binary cross-entropy loss and Dice loss.

\begin{equation}
  \mathcal{L} = \mathcal{L}_{\mathrm{BCE}}(\sigma(\hat{L}_{t,n}), M^L_{t,n}) + \mathcal{L}_{\mathrm{Dice}}(\sigma(\hat{L}_{t,n}), M^L_{t,n})
\end{equation}
Adam was used for optimization, and training required approximately 2~h.

\section{EXPERIMENTAL RESULTS}
\subsection{Experimental Condition}
\begin{figure}[t]
  \centering
  \includegraphics[width=1.0\linewidth]{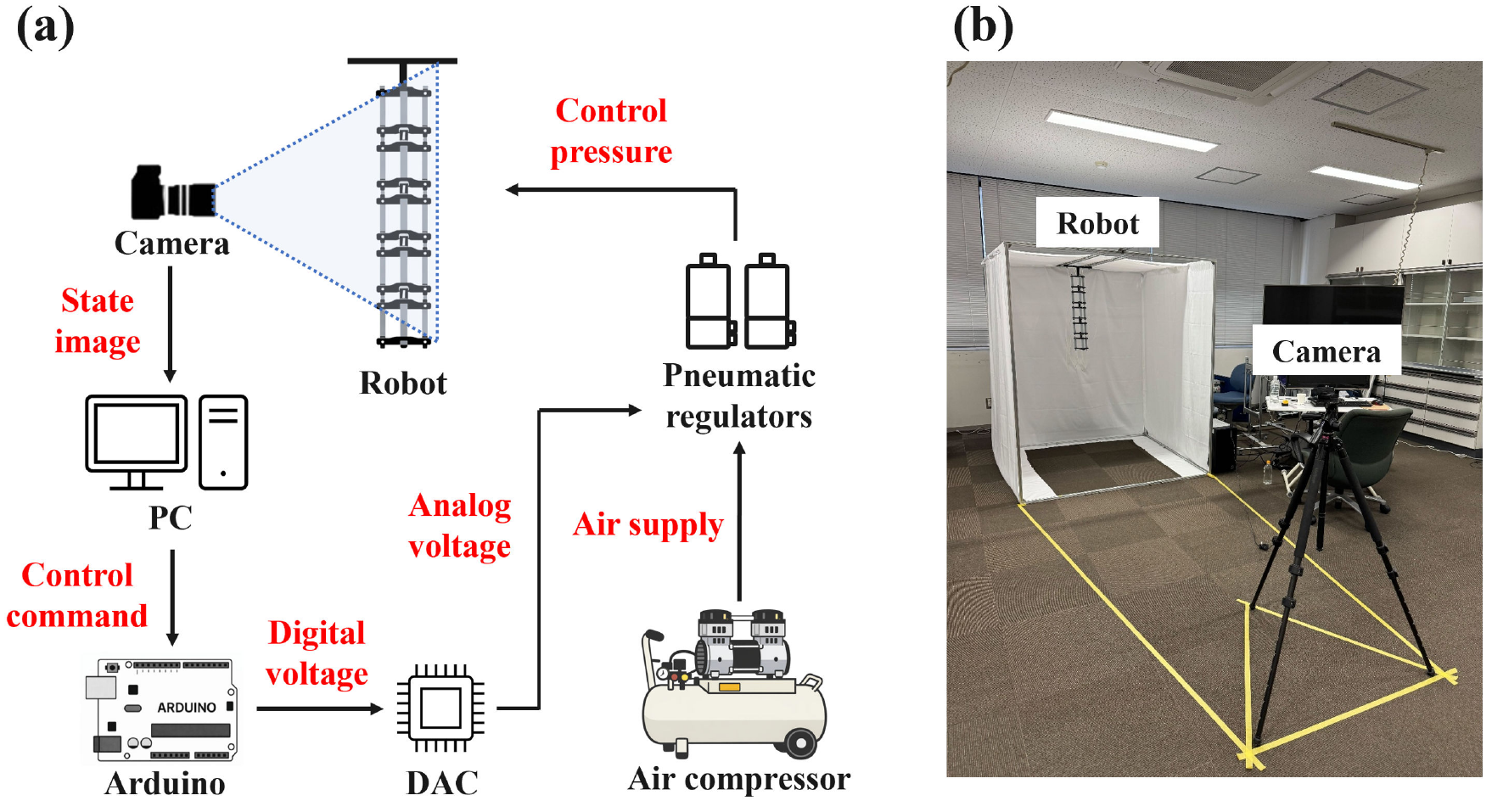}
  \caption{Experimental setup. (a) Control system. The shape of the soft robot is captured, and the pneumatic pressure applied to each actuator is determined based on the observed shape. (b) Experimental setup. The distance between the soft robot and the camera was 3.3~m.}
  \label{fig:system}
\end{figure}

The experimental setup used in this study is shown in Fig.~\ref{fig:system}. The soft robot was suspended from an aluminum frame. Each pneumatic artificial muscle actuator was controlled independently using an electro-pneumatic regulator (ITV0050-3S, SMC), with a maximum operating pressure of 500~kPa. Control signals for the regulators were generated using an Arduino Mega 2560 (Arduino) and two digital-to-analog converters (LTC1660CN, Analog Devices). The robot state was captured using a webcam with a resolution of $1920 \times 1080$ pixels (C920S, Logitech).

Considering the transient oscillation of the robot after pressure application and the quasi-static nature of the posture-control task, the control frequency was set to 0.2~Hz. Achieving a higher control frequency would require a model that explicitly accounts for actuator transient dynamics and dynamically schedules the timing of subsequent commands. Data processing was performed on a PC equipped with an NVIDIA GeForce RTX 3070 GPU and an Intel Core i7-10700F CPU. To reduce the influence of indoor lighting variations, the aluminum frame was enclosed with a white cloth.

Because both the target and current shapes were represented as RGB images, the mean squared error (MSE) computed directly between these images was adopted as the primary evaluation metric.
This choice is consistent with previous studies on image-based shape control of soft robots. However, because image-space MSE does not fully capture all aspects of shape discrepancy, we additionally evaluated overlaid target and actual shapes, the temporal evolution of the error, and convergence behavior under disturbance conditions~\cite{Elijah2023, Marques2024}.

This study does not address the problem of generating arbitrary target images from sketches or high-level planners. Instead, under the assumption that a physically feasible visual target state is provided, we investigate whether a visual controller trained on a single module can be reused for robots with different numbers of modules without additional controller training.

\subsection{Fundamental Characteristics}
\subsubsection{Structural Reconfiguration Experiment}
\begin{figure*}[t]
    \includegraphics[width=0.94\textwidth]{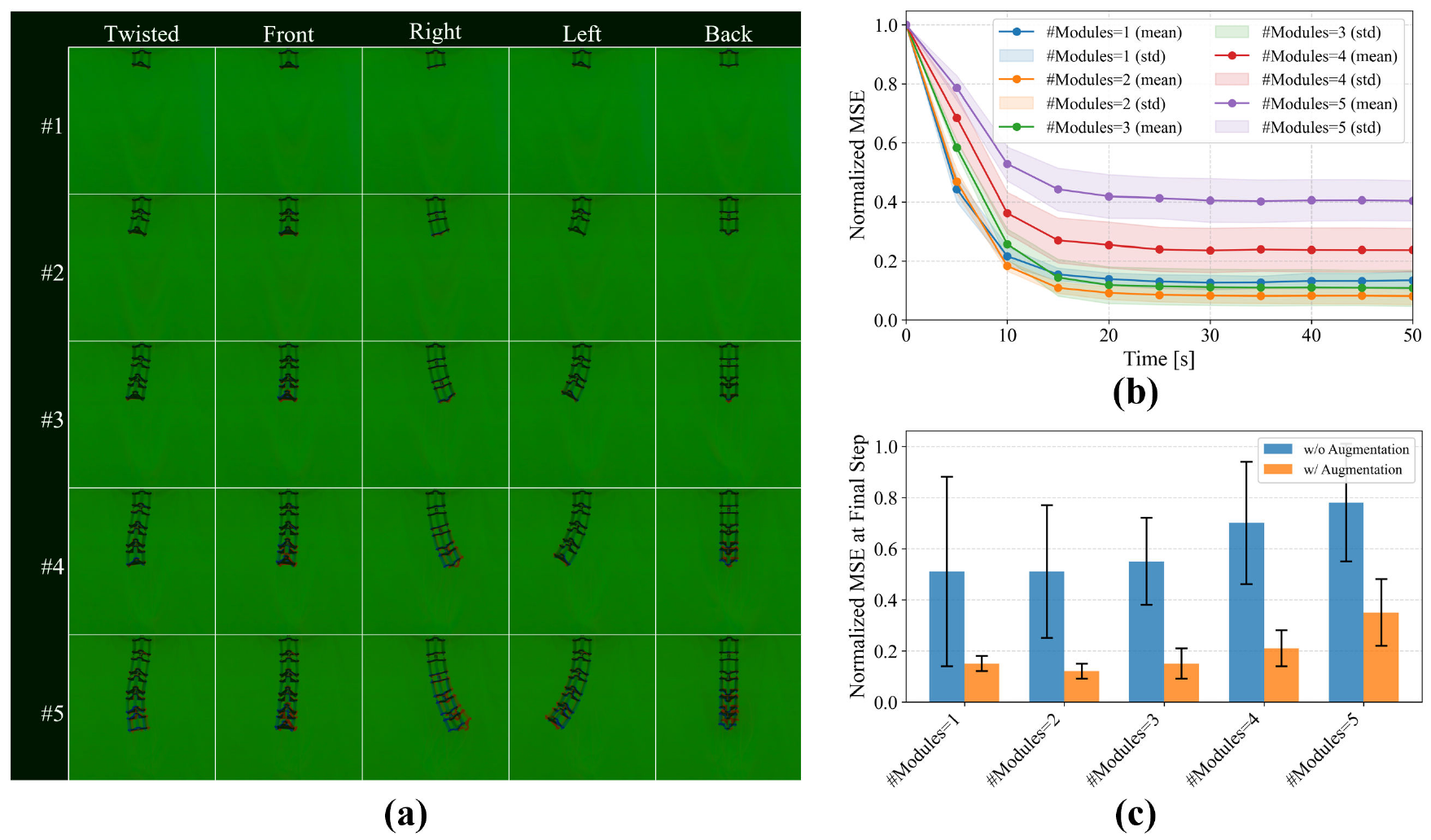}    
    \centering
    \caption{The results confirm that a soft robot with multiple modules can be controlled using a module controller trained only on a single module. (a) The target posture is shown in red and the actual posture in blue. The actual posture converged toward the target posture. (b) Temporal change in the image error between the actual and target postures. For all tested module configurations, the image error was substantially reduced from its initial value and generally converged within approximately 20~s. The shaded regions indicate the standard deviation across the five target postures. (c) Final-step image error with and without data augmentation. The lower image error obtained with data augmentation confirms its importance in training the module controller.}
    \label{fig:fundamental-evaluation}
\end{figure*}
We investigated whether a control model trained on a single module could be extended to robots with multi-module configurations. For each module configuration, five target postures—twisted, forward, rightward, leftward, and backward—were provided as target images, and the ability of the controller to reproduce these postures was evaluated. The controller was trained exclusively on single-module voltage–mask data. None of the target images from the multi-module configurations were included in the training dataset. During closed-loop control, the actuation commands used to generate the target posture were not provided to the controller.

The mean and standard deviation of the MSE were calculated across five target postures. Fig.~\ref{fig:fundamental-evaluation}(a) shows overlays of the target images and the images obtained after the final feedback step. Fig.~\ref{fig:fundamental-evaluation}(b) shows the temporal evolution of the error between the target and actual postures for each module configuration, expressed as the normalized MSE, defined as the MSE at each step divided by the MSE at the initial step. 

As shown in Fig.~\ref{fig:fundamental-evaluation}(a), the robot converged toward shapes close to the target postures for all module configurations. Fig.~\ref{fig:fundamental-evaluation}(b) further shows that the MSE generally decreased and converged over time. 
In configurations with larger numbers of modules, both the mean error and its standard deviation increased compared with those of shorter configurations, indicating that control errors accumulated as the number of modules increased. The image error decreased for all five target postures. 
To evaluate trial-to-trial repeatability, the five-module robot was controlled toward the same twisted target posture in five independent trials. The image error decreased in all trials, and the final normalized MSE was 0.33±0.01. The small inter-trial variation showed low trial-to-trial variability for the tested posture.


We also evaluated the effect of the image data augmentation introduced to emulate the apparent displacement of downstream modules. Fig.~\ref{fig:fundamental-evaluation}(c) shows the mean and standard deviation of the MSE between the target image and the image obtained after the final step for each module configuration, both with and without data augmentation. As shown in Fig.~\ref{fig:fundamental-evaluation}(c), the data augmentation suppressed the increase in error associated with increasing numbers of modules.

\subsubsection{Visual Perturbation Experiments}
To evaluate the robustness of the control model to changes in indoor environmental conditions, three experiments were conducted. First, the illumination of the experimental environment was altered to assess robustness to changes in image brightness. The mean pixel intensity of the camera image decreased from 137.65 under the normal lighting condition to 125.25 under the altered lighting condition.

Second, an obstacle was placed within the camera’s field of view to evaluate robustness to changes in the background. Finally, an affine transformation was applied to the feedback image to assess robustness to geometric variations in the input image. The affine transformation consisted of translations of 15~\text{px} in the horizontal direction and 25~\text{px} in the vertical direction, together with a rotation of $30^\circ$.

\begin{figure}[t]
  \centering
  \includegraphics[width=1.0\linewidth]{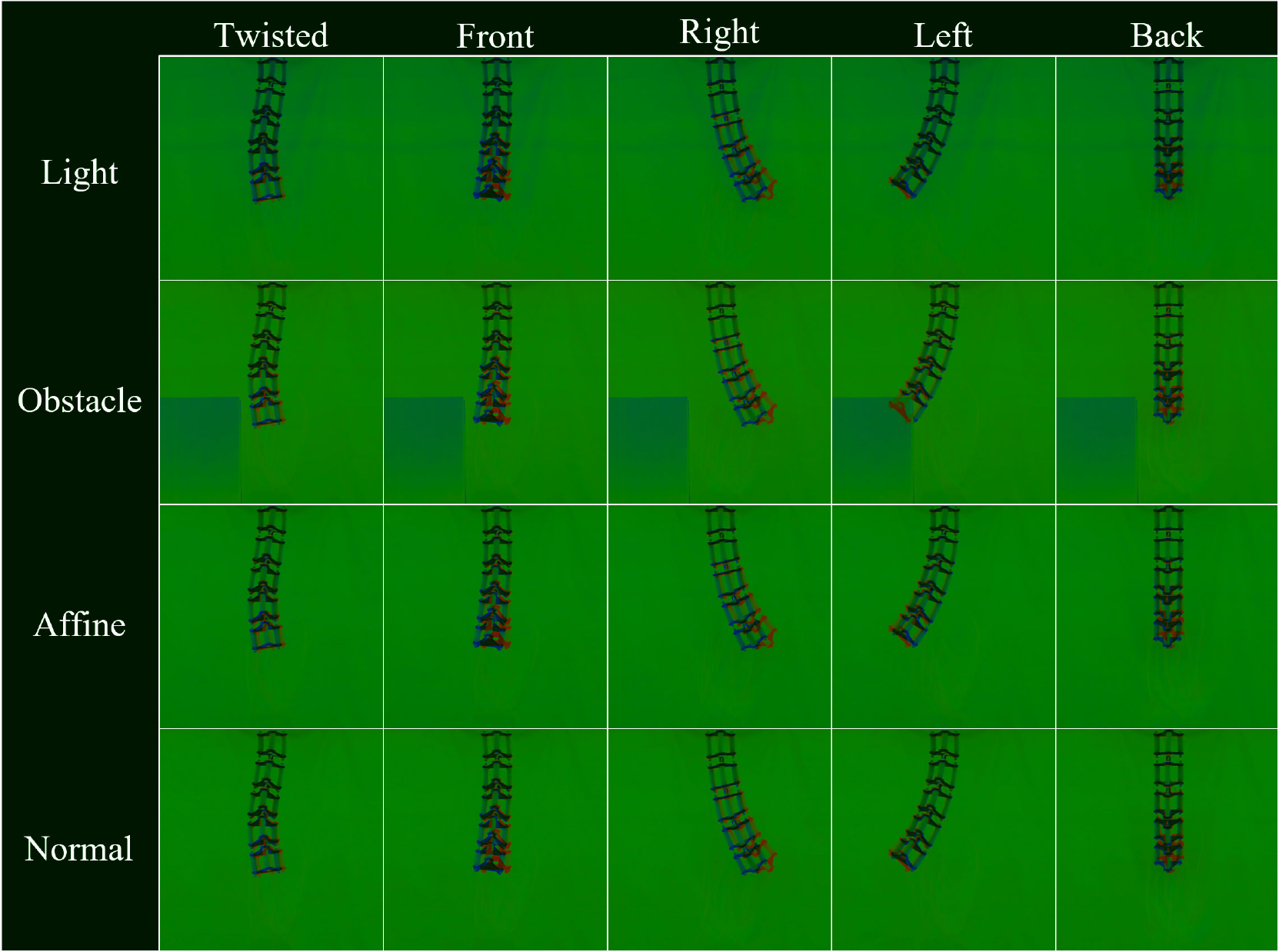}
  \caption{Results of environmental-variation experiments simulating indoor conditions. The target posture is shown in red and the actual posture in blue. The proposed method enabled the actual posture to converge toward the target posture despite changes in illumination, the placement of nearby objects, and synthetic image-plane translation and rotation.}
  \label{fig:assort}
\end{figure}

The five-module robot, which was expected to be most susceptible to environmental variations among the fabricated configurations, was used in all three experiments. Fig.~\ref{fig:assort} shows overlays of the target images and the images obtained after the final step for each condition. In all experiments, the error converged to a level comparable to that observed under the normal condition, demonstrating the robustness to environmental variations.

\subsection{Payload Loading Experiment}
To evaluate the behavior of the transferred controller under an unmodeled loading condition, we assessed its control performance with a payload attached to the robot. The five-module configuration was used in this experiment, and a 500-g mass was attached to its tip.

\begin{figure}[t]
  \centering
  \includegraphics[width=1.0\linewidth]{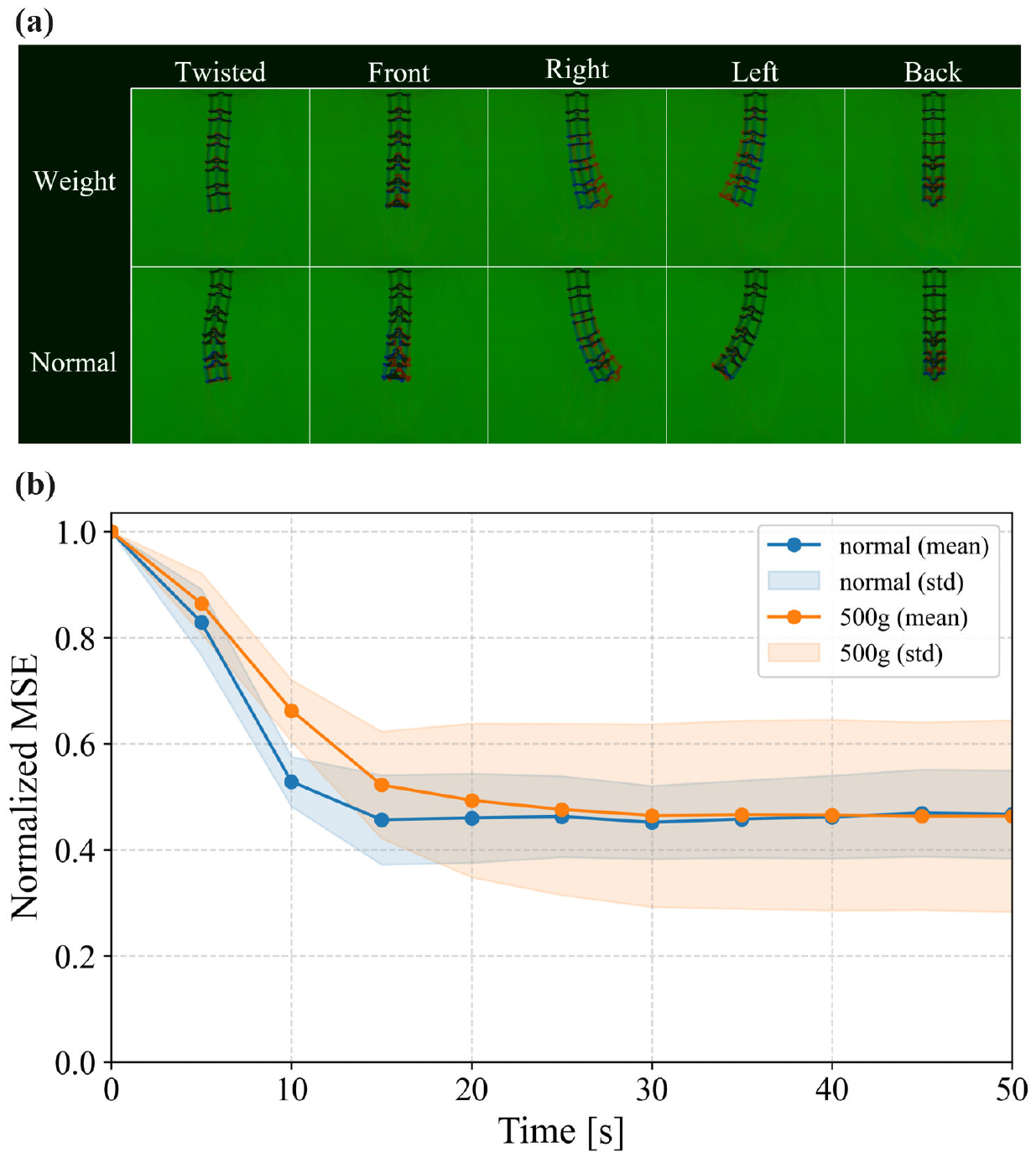}
  \caption{The robot converged to a posture close to the target even with an attached payload. The target posture is shown in red and the actual posture in blue. (a) Comparison between the target and actual postures. (b) Temporal change in the image error between the target and actual postures. The image error converged within approximately 20~s. The shaded regions indicate the standard deviation across the five target postures.}
  \label{fig:weight_mse}
\end{figure}

Fig.~\ref{fig:weight_mse} shows the results of the payload experiment across five target postures. As shown in Fig.~\ref{fig:weight_mse}(a), the magnitude of the visual error varied depending on the target posture. Fig.~\ref{fig:weight_mse}(b) further shows that the standard deviation of the image error across the target postures increased, whereas the mean final image error remained comparable to that obtained without the payload. Although the payload altered the target-dependent control behavior and increased the variation across target postures, visual feedback enabled the robot to converge to a shape close to the target.



\subsection{Mask-Loss Experiment}
To isolate the effect of mask reconstruction, we conducted a controlled mask-dropout experiment in which one channel of the estimated three-channel mask was synthetically removed during the control.
This experiment does not reproduce the full image-level effects of physical occlusion. Instead, it evaluates whether the reconstruction network can prevent control failure when a required mask channel becomes unavailable.

\begin{figure*}[t]
    \includegraphics[width=1.0\textwidth]{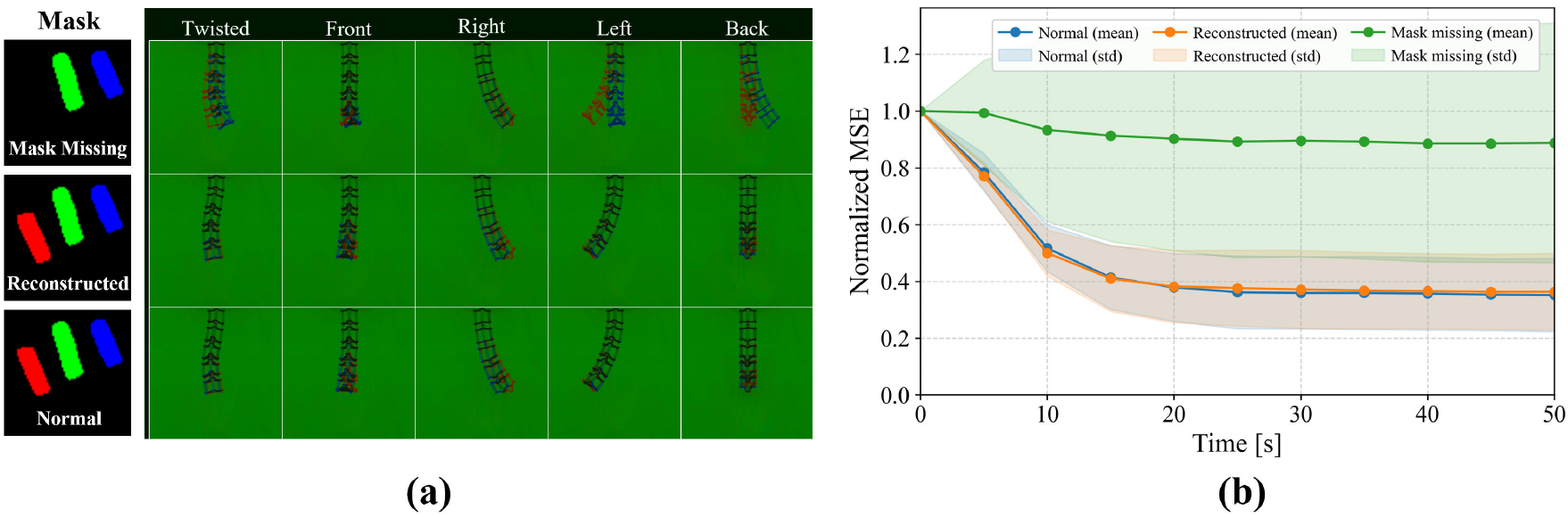}    
    \centering
    \caption{The results confirm that the missing mask could be reconstructed and that the soft robot could still be controlled. (a) The target posture is shown in red and the actual posture in blue. When the mask was missing, the robot hardly moved and did not reach the target posture. In contrast, when the mask was reconstructed, the actual posture converged toward the target posture. (b) Temporal change in the image error between the actual and target postures. When mask reconstruction was applied, the image error decreased over time. The shaded regions indicate the standard deviation across the five target postures.}
    \label{fig:mask}
\end{figure*}

Fig.~\ref{fig:mask}(a), left, shows the mask reconstruction results, while Fig.~\ref{fig:mask}(a), right, shows overlays of the target images and the images obtained after the final feedback step. As shown in Fig.~\ref{fig:mask}(a), the reconstruction network successfully reconstructed the missing mask. In addition, Fig.~\ref{fig:mask}(b) shows that tracking performance comparable to that under normal conditions was maintained.

The results demonstrate robustness against synthetic single-channel mask loss. In real environments, partial occlusion caused by the robot itself, human operators, or manipulated objects can occur frequently. The present experiment represents a simplified failure condition corresponding to the complete loss of one mask channel.

\section{DISCUSSION}
The results demonstrate the feasibility and limitations of reusing
single-module visual control knowledge across serial multi-module
configurations.

\subsection{Scalability of Single-Module Learning}

The objective of this study was not to maximize accuracy for a specific configuration, but to determine whether control knowledge learned from a single module could be reused in larger configurations. The successful tracking of target postures as the number of modules increased supports the effectiveness of this approach. Even for the five-module robot, the final image error was reduced to approximately 40\% of its initial value, demonstrating convergence despite the increased structural length.

For the one- to three-module configurations, small errors were obtained across all target postures, confirming that the single-module controller could be transferred without additional controller training. This supports the central hypothesis that local control knowledge can be composed into whole-body control by applying the same visual controller independently to each module patch.

Geometric data augmentation using rotation and translation was important for this transfer. By approximately reproducing the positional and orientational variations of downstream modules during training, the voltage prediction model became less dependent on the total module number. Thus, generalization depends not only on the controller architecture but also on the geometric variations represented in the single-module training data.

\subsection{Factors Limiting Longer Configurations}

The steady-state error increased for the four- and five-module configurations, although the overall target shapes were still reproduced. This result identifies two main factors limiting transfer to longer robots: changes in load distribution and the loss of depth information in monocular images.

First, increasing the number of modules increases the load and gravitational moment acting on the actuators near the base. This produces a discrepancy between the pressure--deformation relationship learned from a single module and the actual behavior of actuators in longer configurations.

Second, a single two-dimensional view cannot fully distinguish three-dimensional deformation. Motion along the depth direction changes the apparent size and centroid of the artificial-muscle masks independently of actual muscle contraction. Because the controller uses mask shape, size, and position to estimate voltage, such monocular scale changes may induce inappropriate corrections.

Residual errors are also common in image-based soft-robot control because of hysteresis, material nonlinearities, and visual measurement errors~\cite{Elijah2023}. The observed increase in error therefore reflects limitations shared by monocular control of multi-actuator soft robots, rather than a failure specific to the proposed method.

\subsection{Practical Implications and Future Extensions}

Several extensions are possible. First, the visual representation should be generalized beyond actuator-specific masks. Morphology-independent features, such as the robot silhouette or centerline, may allow application to other soft-robot designs.

Second, this study assumes that a physically feasible target image is given. Future work may combine the proposed low-level controller with sketch-based interfaces, geometric models, high-level planners, or generative models for target-image generation.

Third, load compensation may improve performance in longer configurations. Estimating the gravitational torque acting on each module from the detected module number and robot weight, and incorporating it into the control loop, could reduce the accumulated steady-state error without requiring additional control data.

Finally, the current data augmentation is limited to two-dimensional rotation and translation and therefore does not reproduce apparent shortening or scale changes caused by depth motion~\cite{Shentu2024}. Viewpoint-aware augmentation based on Neural Radiance Fields or 3D Gaussian Splatting could synthesize novel-view images and provide training data that better represent three-dimensional visual variations~\cite{NeRF,3DGS}.

\section{CONCLUSION}
This paper presented a module-number-adaptive visual controller for serial modular soft pneumatic robots. A controller trained on single-module data was transferred to one- to five-module configurations by decomposing whole-body images into local module patches. Experiments demonstrated shape-error reduction across all tested configurations, improved transfer through geometric data augmentation, and robustness to environmental changes, payload loading, and synthetic loss of one actuator-mask channel. These results support scalable, markerless whole-body control without configuration-specific controller training. 


\bibliographystyle{IEEEtran}
\bibliography{IEEE_RA-L}

@article{tang2026review,
  title={A review of modular soft robots: Drive modes, motion modes, and connection modes},
  author={Tang, Dedong and Chen, Xingyu and Zhang, Yongde},
  journal={Proceedings of the Institution of Mechanical Engineers, Part C: Journal of Mechanical Engineering Science},
  volume={240},
  number={10},
  pages={3650--3676},
  year={2026},
  publisher={SAGE Publications Sage UK: London, England}
}

@INPROCEEDINGS{shariati2021,
  author={Shariati, Azadeh and Shi, Jialei and Spurgeon, Sarah and Wurdemann, Helge A},
  booktitle={2021 IEEE/RSJ International Conference on Intelligent Robots and Systems (IROS)}, 
  title={Dynamic modelling and visco-elastic parameter identification of a fibre-reinforced soft fluidic elastomer manipulator}, 
  year={2021},
  volume={},
  number={},
  pages={661-667},
  doi={10.1109/IROS51168.2021.9636785}}

@ARTICLE{Ferrentino2022,
  author={Ferrentino, Pasquale and López-Díaz, Antonio and Terryn, Seppe and Legrand, Julie and Brancart, Joost and Van Assche, Guy and Vázquez, Ester and Vázquez, Andrés and Vanderborght, Bram},
  journal={IEEE Robotics and Automation Letters}, 
  title={Quasi-Static FEA Model for a Multi-Material Soft Pneumatic Actuator in SOFA}, 
  year={2022},
  volume={7},
  number={3},
  pages={7391-7398},
  doi={10.1109/LRA.2022.3183254}}

@article{Webster2010,
  author={Webster III, Robert J. and Jones, Bryan A.},
  title={Design and kinematic modeling of constant curvature continuum robots: A review},
  journal={The International Journal of Robotics Research},
  volume={29}, number={13}, pages={1661--1683}, year={2010}
}

@article{DellaSantina2020,
  author={Della Santina, Cosimo and Bicchi, Antonio and Rus, Daniela},
  title={On an improved state parametrization for soft robots with piecewise constant curvature and its use in model based control},
  journal={IEEE Robotics and Automation Letters},
  volume={5}, number={2}, pages={1001--1008}, year={2020}
}

@ARTICLE{DellaSantinaSurvey2023,
  author={Della Santina, Cosimo and Duriez, Christian and Rus, Daniela},
  journal={IEEE Control Systems}, 
  title={Model Based Control of Soft Robots: A Survey of the State of the Art and Open Challenges}, 
  year={2023},
  volume={43},
  number={3},
  pages={30-65},
  doi={10.1109/MCS.2023.3253419}}

@article{Till2019,
  author={Till, John and Aloi, Vincent and Rucker, Caleb},
  title={Real-time dynamics of soft and continuum robots based on Cosserat rod models},
  journal={The International Journal of Robotics Research},
  volume={38}, number={6}, pages={723--746}, year={2019}
}

@INPROCEEDINGS{Largilliere2015,
  author={Largilliere, Frederick and Verona, Valerian and Coevoet, Eulalie and Sanz-Lopez, Mario and Dequidt, Jeremie and Duriez, Christian},
  booktitle={2015 IEEE International Conference on Robotics and Automation (ICRA)}, 
  title={Real-time control of soft-robots using asynchronous finite element modeling}, 
  year={2015},
  volume={},
  number={},
  pages={2550-2555},
  doi={10.1109/ICRA.2015.7139541}}

@INPROCEEDINGS{Tonkens2021,
  author={Tonkens, Sander and Lorenzetti, Joseph and Pavone, Marco},
  booktitle={2021 IEEE International Conference on Robotics and Automation (ICRA)}, 
  title={Soft Robot Optimal Control Via Reduced Order Finite Element Models}, 
  year={2021},
  volume={},
  number={},
  pages={12010-12016},
  doi={10.1109/ICRA48506.2021.9560999}}

@ARTICLE{Chen2025,
  author={Chen, Zixi and Guan, Qinghua and Hughes, Josie and Menciassi, Arianna and Stefanini, Cesare},
  journal={IEEE Transactions on Robotics}, 
  title={A Versatile Neural Network Configuration Space Planning and Control Strategy for Modular Soft Robot Arms}, 
  year={2025},
  volume={41},
  number={},
  pages={4269-4282},
  doi={10.1109/TRO.2025.3582807}}

@article{ChinMajidi2020,
  author={Chin, Kevin and Hellebrekers, Tess and Majidi, Carmel},
  title={Machine learning for soft robotic sensing and control},
  journal={Advanced Intelligent Systems},
  volume={2}, number={6}, pages={1900171}, year={2020}
}

@ARTICLE{ABCD,
  author={Krauss, Henrik and Licher, Johann and Takeishi, Naoya and Raatz, Annika and Yairi, Takehisa},
  journal={IEEE Robotics and Automation Letters}, 
  title={Learning Visually Interpretable Oscillator Networks for Soft Continuum Robots From Video}, 
  year={2026},
  volume={11},
  number={8},
  pages={9495-9502},
  doi={10.1109/LRA.2026.3703241}}

@ARTICLE{Elijah2023,
  author={Almanzor, Elijah and Ye, Fan and Shi, Jialei and Thuruthel, Thomas George and Wurdemann, Helge A. and Iida, Fumiya},
  journal={IEEE Transactions on Robotics}, 
  title={Static Shape Control of Soft Continuum Robots Using Deep Visual Inverse Kinematic Models}, 
  year={2023},
  volume={39},
  number={4},
  pages={2973-2988},
  doi={10.1109/TRO.2023.3275375}}

@ARTICLE{Marques2024,
AUTHOR={Marques Monteiro, Richard  and Shi, Jialei  and Wurdemann, Helge  and Iida, Fumiya  and George Thuruthel, Thomas },
TITLE={Visuo-dynamic self-modelling of soft robotic systems},
JOURNAL={Frontiers in Robotics and AI},
VOLUME={11},
YEAR={2024},
DOI={10.3389/frobt.2024.1403733},
}

@misc{gandhi20263dshapecontrolextensible,
  author  = {Abhinav Gandhi and Shou-Shan Chiang and
             Cagdas D. Onal and Berk Calli},
  title   = {{3D} Shape Control of Extensible Multi-Section
             Soft Continuum Robots via Visual Servoing},
  journal = {arXiv preprint arXiv:2602.19273},
  year    = {2026}
}

@article{usui2021,
  author={Usui, T and Ishizuka, H and Kawasetsu, T and Hosoda, K and Ikeda, S and Oshiro, O},
  journal={Sensors and Actuators A: Physical}, 
  title={Soft capacitive tactile sensor using displacement of air-water interface }, 
  year={2021},
  volume={332},
  number={1},
  pages={113133},
  }

@INPROCEEDINGS{soter2019,
  author={Soter, Gabor and Garrad, Martin and Conn, Andrew T. and Hauser, Helmut and Rossiter, Jonathan},
  booktitle={Proceedings of 2019 2nd IEEE International Conference on Soft Robotics (RoboSoft)}, 
  title={Skinflow: A soft robotic skin based on fluidic transmission}, 
  year={2019},
  volume={},
  number={},
  pages={355-360},
  doi={10.1109/ROBOSOFT.2019.8722744}}

@article{chen2024,
  author  = {Zixi Chen and Matteo Bernabei and Vanessa Mainardi
             and Xuyang Ren and Gastone Ciuti and Cesare Stefanini},
  title   = {A Novel and Accurate {BiLSTM} Configuration Controller
             for Modular Soft Robots with Module Number Adaptability},
  journal = {Soft Robotics},
  year    = {2024},
  month   = dec,
  note    = {Early access},
  doi     = {10.1089/soro.2024.0015}
}

@misc{recipe,
  author = {Yoichi Masuda},
  title = {Recipe for {McKibben}-Type Pneumatic Muscle},
  howpublished = {\url{https://ishikawa-lab.sakura.ne.jp/mckibben_eng}},
    note = {Accessed: 2026-07-30}
}

@misc{wu2019detectron2,
author = {Wu, Yuxin and Kirillov, Alexander and Massa, Francisco and Lo, Wan-Yen and Girshick, Ross},
title = {Detectron2},
url = {https://github.com/facebookresearch/detectron2},
  note = {Accessed: 2026-07-30}
}

@misc{wada2021labelme,
  author = {Wada, Kentaro},
  title = {Labelme: Image Polygonal Annotation with Python},
  year = {2021},
  url = {https://github.com/wkentaro/labelme},
  note = {Accessed: 2026-07-30}
}

@InProceedings{NeRF,
author={Mildenhall, Ben
and Srinivasan, Pratul P.
and Tancik, Matthew
and Barron, Jonathan T.
and Ramamoorthi, Ravi
and Ng, Ren},
title={{NeRF}: Representing Scenes as Neural Radiance Fields for View Synthesis},
booktitle={Computer Vision -- ECCV 2020},
year={2020},
address={Cham},
pages={405--421},
}

@Article{3DGS,
      author       = {Kerbl, Bernhard and Kopanas, Georgios and Leimk{\"u}hler, Thomas and Drettakis, George},
      title        = {{3D} Gaussian Splatting for Real-Time Radiance Field Rendering},
      journal      = {ACM Transactions on Graphics},
      number       = {4},
      volume       = {42},
      month        = {July},
      year         = {2023}
}

@ARTICLE{Kosaka2025,
  author={Kosaka, Shota and Kimura, Kentaro and Yamamoto, Seiichi and Ishizuka, Hiroki and Masuda, Yoichi and Punpongsanon, Parinya and Ikeda, Sei and Oshiro, Osamu},
  journal={IEEE Robotics and Automation Letters}, 
  title={Reconfigurable Soft Pneumatic Actuators Using Multi-Material Self-Healing Polymers}, 
  year={2025},
  volume={10},
  number={5},
  pages={4938-4945},
  doi={10.1109/LRA.2025.3554971}}

@ARTICLE{Shentu2024,
  author={Shentu, Chengnan and Li, Enxu and Chen, Chaojun and Dewi, Puspita T. and Lindell, David B. and Burgner-Kahrs, Jessica},
  journal={IEEE Robotics and Automation Letters}, 
  title={{MoSS}: Monocular Shape Sensing for Continuum Robots}, 
  year={2024},
  volume={9},
  number={2},
  pages={1524-1531},
  doi={10.1109/LRA.2023.3346271}}
\end{document}